\documentclass{article}
\usepackage{graphicx} 

\usepackage{natbib}
\setcitestyle{authoryear}
\usepackage[colorlinks=true, linkcolor=blue, citecolor=blue, urlcolor=blue]{hyperref}
\usepackage{booktabs}
\usepackage{multirow}
\usepackage{amsmath}
\DeclareMathOperator{\argmax}{argmax}
\usepackage{orcidlink}

\usepackage{xcolor}
\usepackage{todonotes}

\title{Beyond Off-the-Shelf Models: A Lightweight and Accessible Machine Learning Pipeline for Ecologists Working with Image Data}
\author{Clare Chemery$^a$ \orcidlink{0009-0006-0710-172X}, Hendrik Edelhoff$^c$ \orcidlink{0000-0001-6354-8952}, Ludwig Bothmann$^{a,b,*}$ \orcidlink{0000-0002-1471-6582}\\[2ex]
\small
$^a$Department of Statistics, LMU Munich, Germany\\
\small$^b$Munich Center for Machine Learning (MCML), Germany\\
\small$^c$Research Unit Wildlife Biology and Management,\\
\small Bavarian State Institute for Forestry, Freising, Germany\\[2ex]
\small$^*$Corresponding author: \texttt{ludwig.bothmann@lmu.de}\\[2ex]
\normalsize
}
\begin{document}

\maketitle

\begin{abstract}
We introduce a lightweight experimentation pipeline designed to lower the barrier for applying machine learning (ML) methods for classifying images in ecological research. We enable ecologists to experiment with ML models independently, thus they can move beyond off-the-shelf models and generate insights tailored to local datasets and specific classification tasks and target variables. Our tool combines a simple command-line interface for preprocessing, training, and evaluation with a graphical interface for annotation, error analysis, and model comparison. This design enables ecologists to build and iterate on compact, task-specific classifiers without requiring advanced ML expertise. As a proof of concept, we apply the pipeline to classify red deer (\textit{Cervus elaphus}) by age and sex from 3392 camera trap images collected in the Veldenstein Forest, Germany. Using 4352 cropped images containing individual deer labeled by experts, we trained and evaluated multiple backbone architectures with a wide variety of parameters and data augmentation strategies. Our best-performing models achieved 90.77\% accuracy for age classification and 96.15\% for sex classification. These results demonstrate that reliable demographic classification is feasible even with limited data to answer narrow, well-defined ecological problems. More broadly, the framework provides ecologists with an accessible tool for developing ML models tailored to specific research questions, paving the way for broader adoption of ML in wildlife monitoring and demographic analysis.
\end{abstract}

\section{Introduction}

From diagnosing diseases to writing code, ML has reshaped entire industries \citep{houck_space_2025, brynjolfsson_what_2018}, but its potential in ecology remains underutilized relative to the volume of data created in the field. Camera trap images offer an enormous opportunity for automated wildlife monitoring, but adoption has been slowed by the technical challenges of implementing ML models and the scarcity of large, heterogeneous annotated datasets \citep{oeser_integrating_2023, tabak_cameratrapdetector_2022, tuia_perspectives_2022}. These challenges have led to the adoption of task-specific models trained for specific regions and species. While computationally efficient, these models often falter when applied to new species or habitats \citep{schneider_three_2020}, leading to high technical overhead as a new model must be trained for each new task. As one solution, an active-learning pipeline was introduced by \citet{bothmann_automated_2023} that allows ecology researchers to iteratively train a species classifier while efficiently selecting the next batch of images to be annotated to maximize the benefit to model performance.

Recent research in foundation models has led to the introduction of mixture of experts models like Mistral's Mixtral 8x7B model \citep{jiang_mixtral_2024}. These models are composed of a router that selects the relevant blocks from an array of task-specific experts to work across a variety of domains. Using this conceptual framework, we propose an easy-to-use tool that allows ecology researchers to train task-specific experts that they can call on to perform tasks in a variety of ML problems.

In this work, we present a flexible, easy-to-use pipeline for building many compact models at once for specific image classification tasks that simplifies the preprocessing, training, and comparison processes end-to-end. As a proof of concept, we experiment with different model architectures and parameter sets for classifying images of red deer by age and sex from camera trap images. The system offers an intuitive 7-step workflow that guides users from raw camera trap images through MegaDetector wildlife detection, web-based annotation, and model training using simple command-line interfaces with preset configurations.\footnote{All code is available at \url{https://github.com/slds-lmu/wildlife-age-sex} and more information can be found in the read-me file at \url{https://github.com/slds-lmu/wildlife-age-sex/blob/main/README.md}.} The tool's ease of use is demonstrated by a comprehensive demo pipeline that can be run immediately after installation and its web interface that launches with a single command. Our framework enables ecologists with limited ML experience to tailor models to local datasets and new classification tasks via a user interface (UI) for annotation, error analysis and model comparison and a simple command-line interface (CLI) for preprocessing, training, and evaluation.

\section{Related Work}

A few integrated pipeline proposals exist in the ecology field already. \cite{10_24072_pcjournal_565} introduced the Mega-Efficient Wildlife Classification (MEWC) workflow, which leverages the containerization tool Docker to create a modular pipeline. MEWC lets users perform the full range of tasks (object detection, species ID, etc.)\ on local or cloud graphical processing units, and automatically writes classifications into image metadata and CSV files. This pipeline is designed to handle the large amounts of data required to train a general classification model, but for many tasks, this level of computational power is not needed, and a more lightweight framework that allows for quick iteration is preferable.

Other authors have provided software for performing inference using their models as in \cite{conservation_ai} or adapting them to specific tasks as in \cite{tabak_cameratrapdetector_2022} with their R package Machine Learning for Wildlife Image Classification. Other open-source software is available for the baseline task-agnostic portions of an ML workflow, including Camelot \citep{camelot_2017}, for storing camera trap images and their metadata and labeling species, and AddaxAI \citep{addax_ai}, for performing inference and reviewing results from pretrained models. These tools can be very helpful for leveraging existing models as part of a research team's workflow, but we focus on enabling teams to train their own models and experiment with ML solutions to specific questions.

For our proof of concept, age and sex classification with ML is, as yet, an underexplored area of ecology research. Sex classification has been performed on curated datasets of high-quality images among zebra fish \citep{hosseini_efficient_2019}, giant pandas \citep{qi_giant_2022}, and humans \citep{kaur_sex_2025}.  Some studies on age classification have also been performed among human subjects as in \cite{dey_human_2024}. While the literature showed that this nuanced classification could be done with existing ML methods, the images used were of much higher and more consistent quality than images captured by camera traps. Therefore, the problem of age and sex classification from opportunistic image capture is still outstanding in ecology.

Age and sex classification results can be used to estimate the population demographics. The demographics of a species can, in turn, be used to predict population growth and set population control policies like hunting quotas. In conjunction with other population estimation methods like those used by \cite{zampetti_towards_2024}, even learning the approximate distribution, rather than absolute counts, of age and sex demographics may be sufficient for these applications.

The ability to accurately classify the sex and age class of individuals detected in camera trap data is crucial for addressing various ecological questions and enabling deeper, more meaningful analyses. These classification results are essential for estimating population demographics, such as the distribution of different age classes and the adult sex ratio (ASR) \citep{wohlfahrt_red_2026}—the proportion of males within the adult population—which, in turn, informs population management by predicting population growth and guiding population control policies, like setting appropriate hunting quotas; even learning the approximate demographic distribution, when used in conjunction with other population estimation methods \citep[e.g.,][]{zampetti_towards_2024}, may be sufficient. The ASR is a vital metric for estimating population size, recruitment, harvest rate, and natural mortality \citep{solberg_change--sex_2005}, and in ungulates, the ASR influences conception time and calving dates, directly impacting newborn survival \citep{holand_effect_2003}, making its management critical for regulating population size \citep{hovestadt_gender-specific_2014, solberg_biased_2002}; ratios like that of adult females to young \citep{bender_uses_2006, bonenfant_can_2005} offer additional context regarding reproductive success. Beyond basic demographics, the classification of individuals by age and sex allows for more complex, in-depth ecological studies, enabling the calculation of sex-specific or age-class-specific\textbf{ }space use through occupancy models, and facilitating the differentiation of behavioral patterns, such as comparing the spatial utilization or activity budgets of adult animals with young dependents versus those without \citep{thomas_influence_2025}.

\section{Methods}
\subsection{Set Up and Requirements}

The pipeline code can be found in our GitHub repository, \href{https://github.com/slds-lmu/wildlife-age-sex.git}{\texttt{wildlife-age-sex}}.\footnote{\url{https://github.com/slds-lmu/wildlife-age-sex}} The repository includes a comprehensive guide to the code base, a step-by-step tutorial for configuring the user's local environment and implementing the pipeline for a new task, and demo data for testing the pipeline, making it easy for less technical users to leverage this tool for their research.

For larger samples or high-resolution images, the pipeline should be run on a remote server with additional capacity, but for more lightweight data sets (smaller, lower resolution), the pipeline can also be run on a local machine. In order to run the UI components of the pipeline, the machine must have access to port 8501 on the local host, either by default or via SSH port forwarding from the remote server to a local machine, but in general, the tool is operating-system-agnostic, i.e., runs in Windows, Mac OS, and Linux. 

\subsection{Bounding Box Detection and Annotation Interface}

Incorporating PyTorch-Wildlife MegaDetector (MD) \citep{Beery_Efficient_Pipeline_for} before labeling has been shown to improve performance over using raw images \citep{velez_evaluation_2023, beery_efficient_2019}. Our pipeline includes a script that generates bounding box detections from raw images using the MD, which locates each animal and assigns a confidence score to the detection.
 
For our dataset, the bounding box detection and annotation was done by experts with internal software tools over a long period of time resulting in variation in schema (different column names and/or data types used for properties shared across batches), class values (e.g., "sub-adult" vs. "juvenile"), and bounding box definitions between MD-V5 and -V6. 

As part of our work, we introduced a unified bounding box detection and annotation workflow. This workflow minimizes manual transformations in order to maximize the cross-compatibility of datasets generated for different ML tasks or by different researchers. We created a simple script that runs MD-V6 (YOLOv10-e) on a folder of images and defines a consistent output schema for downstream compatibility. For each bounding box detection, an entry is created with a unique bounding box ID, the image ID, the path to the image, the detection category (as generated by the MD, 0 being "animal"), the bounding box coordinates, and the confidence of the bounding box.

\begin{figure}[htbp]
  \centering
  \includegraphics[width=0.95\textwidth]{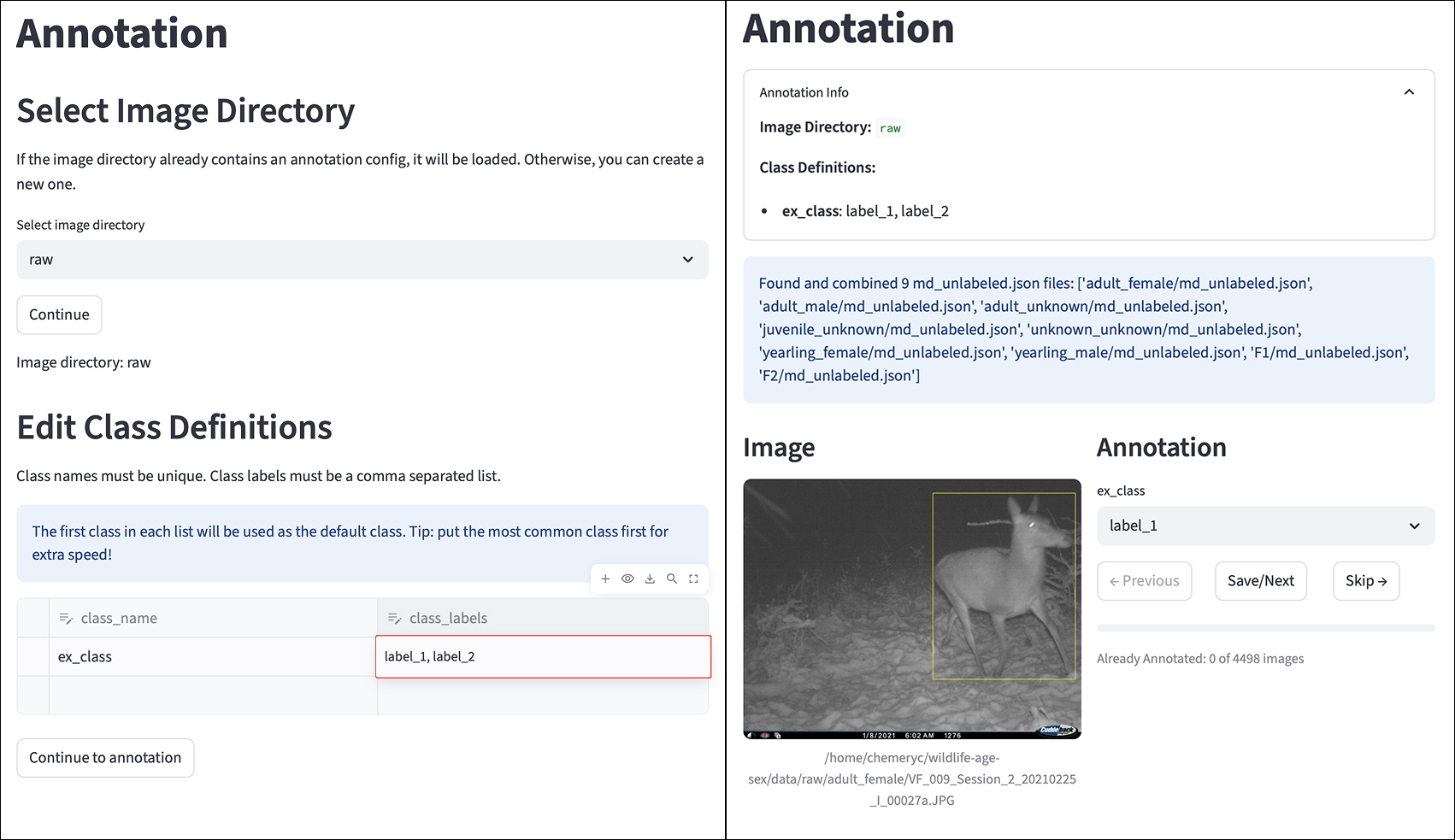}
  \caption{Annotation Interface in Streamlit}
  \label{fig:annotation}
\end{figure}

Once a JSON file of detections is generated using this schema, the user can launch the project’s UI, hosted locally using Streamlit. In the annotation page, users can select the directory that contains the images and MD detections file, define the classes (here, referring to the classification target, e.g., age or sex), and their possible labels (e.g., female, male, unknown), and annotate the bounding boxes using simple drop-down menus (\autoref{fig:annotation}). A JSON file of annotations is created in the image directory, and additional metadata, such as camera location or the time the image was captured, can be added using the enrichment script we provide. These additional metadata can be used for stratified splitting or to segment evaluations later on in the pipeline.

\subsection{Configuration File}

The configuration file is one unified location where all the specifications of an experiment are defined from preprocessing through evaluation. This is a human-readable and -writable file, written in Tom’s Obvious Minimal Language (TOML). 

\begin{figure}[htbp]
  \centering
  \includegraphics[width=0.95\textwidth]{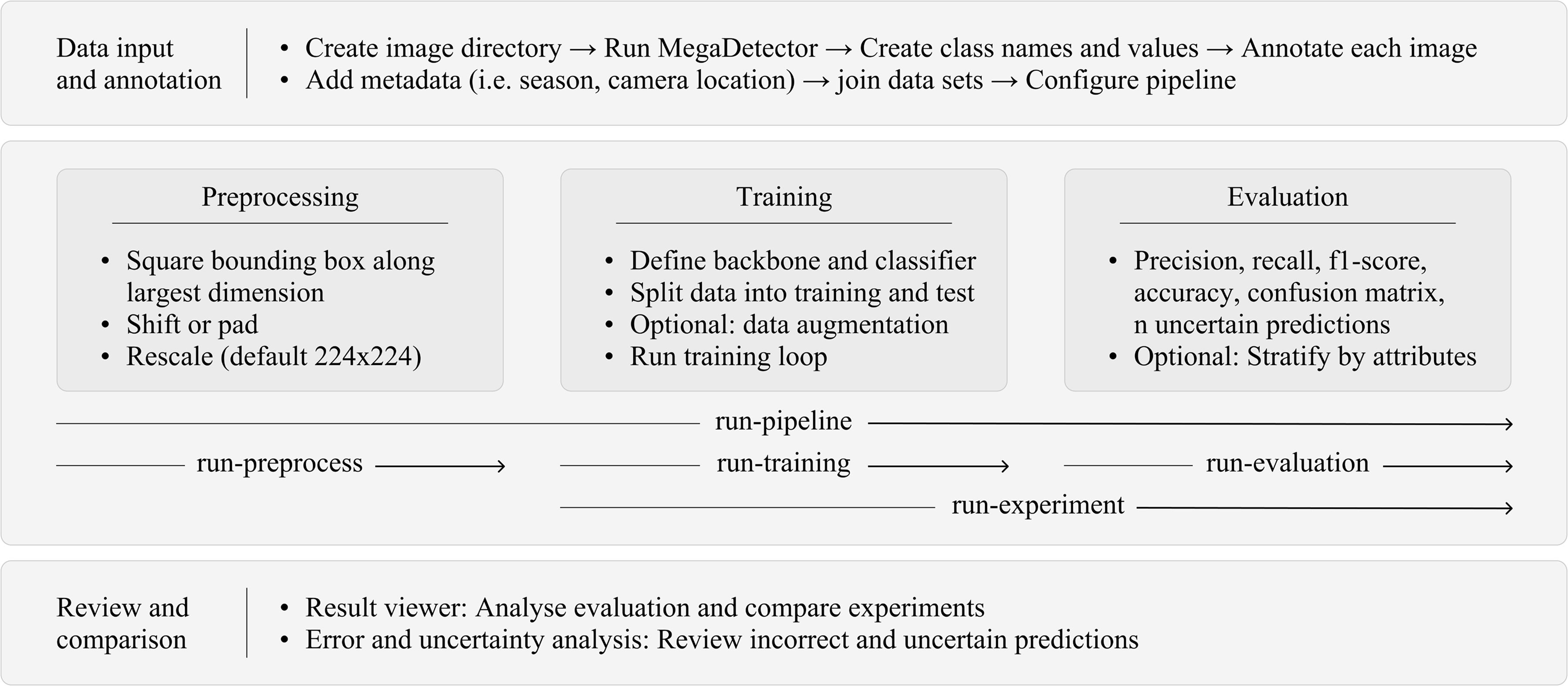}
  \caption{Pipeline for Classification Experiments}
  \label{fig:pipeline}
\end{figure}

The configuration file has a few important sections: IO (input-output) where users define global data and model paths, and sections in which the arguments for individual steps in the pipeline are set, namely for preprocessing, training, and evaluation. The configuration file can be written to accommodate many different classification tasks with different target variables, datasets, and training loops. These files create a record of the data and parameters used to train each model and define experiments within the pipeline.  \autoref{fig:pipeline} provides an overview of the entire pipeline; each step will be explained in the following. 

\subsection{Preprocessing}

As a starting point for the following experimental setup, the dataset is expected to contain the coordinates and the confidence score of the bounding box, the target variable, and any other metadata to be used for stratification of the train-test split or in evaluation.

As the first step in preprocessing, the user-defined confidence threshold (CT) is applied. This filters out bounding boxes that may be empty or contain only small parts of an animal from all further steps in the pipeline. By default, only high-confidence ($\geq 0.96$) bounding boxes are used.

The CT used can be considered an additional parameter for tuning, as in \citet{bothmann_automated_2023}. The optimal value of this parameter depends strongly on the task, the training data available, the MD version, and the confidence of the bounding boxes on which the user would like to run inference. 

Next, the bounding boxes are cropped into a square, from the centroid of the original box using the longest side as the side length, to capture peripheral information such as antlers that may not be captured by a conservative bounding box. We utilize a shifting strategy that translates the bounding box so that it aligns with the edges of the image without decreasing its size. Alternatively, users can choose the traditional padding strategy which adds black pixels where the bounding box exceeds the dimensions of the image.

Finally, the images are rescaled to a user-defined dimension, which for most standard pretrained models is optimally 224 by 224, as we use in our implementation.

\subsection{Model Architecture and Training}

A classification task is defined on a dataset $D=\left((x^{(1)},y^{(1)}),\dots,(x^{(n)},y^{(n)})\right) $ made up of labeled images, where each label $y^{(i)} \in \{1,\dots,g\}$ denotes the class of the corresponding image $x^{(i)}$, with a total of $g$ classes. The objective is to train a classification model $\hat{\pi}(\cdot)$ that maps an input image $x$ to a vector of probability scores $\hat{\pi}(x)$ with elements ${\pi_k\ \in\ \{\pi_1,\ \ldots,\ \pi_g\}}$ corresponding to each class. The predicted hard label is then found from $\hat{\pi}(x)$ as

\begin{equation}
    \hat{y} = \argmax_{k \in \{1, \ldots, g\}} \hat{\pi}_k(x)
\end{equation}

In the training loop, we use PyTorch to construct a custom model comprised of the pretrained backbone model chosen by the user and a classifier configured to match the selected backbone and the number of output features for a given task. Currently, ResNet50 \citep{he_deep_2015},  VGG19 \citep{simonyan_very_2014}, DenseNet161 and DenseNet201 \citep{huang_densely_2016} are supported. 
When splitting the dataset into training and test -- currently an 80/20\% split -- the user can select to stratify on a certain attribute (such as season, year, etc.). We implemented stratification on the target column to preserve the distribution of classes across training and test sets. 

Once the training set is defined, the parameters of the custom model architecture are trained according to the user-specified training parameters. Our well-documented configuration file allows users to review all possible parameters and default values in one place to minimize false assumptions and to surface all potential tuning options. 
One important feature of the training loop is data augmentation, previously shown to improve performance in image classification tasks \citep{shorten_survey_2019}. The user has the option to train without or with light, medium, or strong augmentation. By default, the augmentation includes flips, rotations, brightness and contrast variation, blur, noise, and more. These transformations were selected to mimic variation that is likely to occur in the camera trap setting, such as low lighting or moving subjects.

The training loop is customizable but has two main sections: (i) transfer learning, which tunes the classifier layer only and leverages the general knowledge of pretrained models for the rest of the architecture \citep{kurkova_survey_2018}, and (ii) finetuning, which tunes the classifier layer and deeper layers of the model. Each stage of training has its own set of parameters that can be adjusted to achieve optimal performance. We implemented both stages with an 85\%/15\% training and validation split among the training data for each epoch. 

\subsection{Evaluation}

The evaluation step computes classification metrics, namely precision, recall, F1-score, accuracy, and a confusion matrix. The trained model outputs confidence scores alongside its predictions. Predictions made with a confidence score below a certain CT (default 0.5) are flagged as uncertain and excluded from automated metrics. The user can change this behavior in the configuration file depending on the task they plan to perform with the finalized model.

The user can also select an attribute on which to provide stratified metrics. We investigated the model’s performance on summer data (May through September) versus winter data as many of the key markings of age and sex among red deer (antlers of male deer, white spots of yearlings) are only extant in the summer months. 

The evaluation stage also records the incorrect prediction and their bounding boxes for later review. Whether uncertain images were excluded from the evaluation or not, these images with their predictions are also recorded for review. 

\subsection{Results and Error Viewing}

When running multiple experiments with different parameters and configurations, it is important for users to quickly evaluate and compare model performance. Our Streamlit application provides three pages designed to facilitate this process.

\begin{figure}[htbp]
  \centering
  \includegraphics[width=0.95\textwidth]{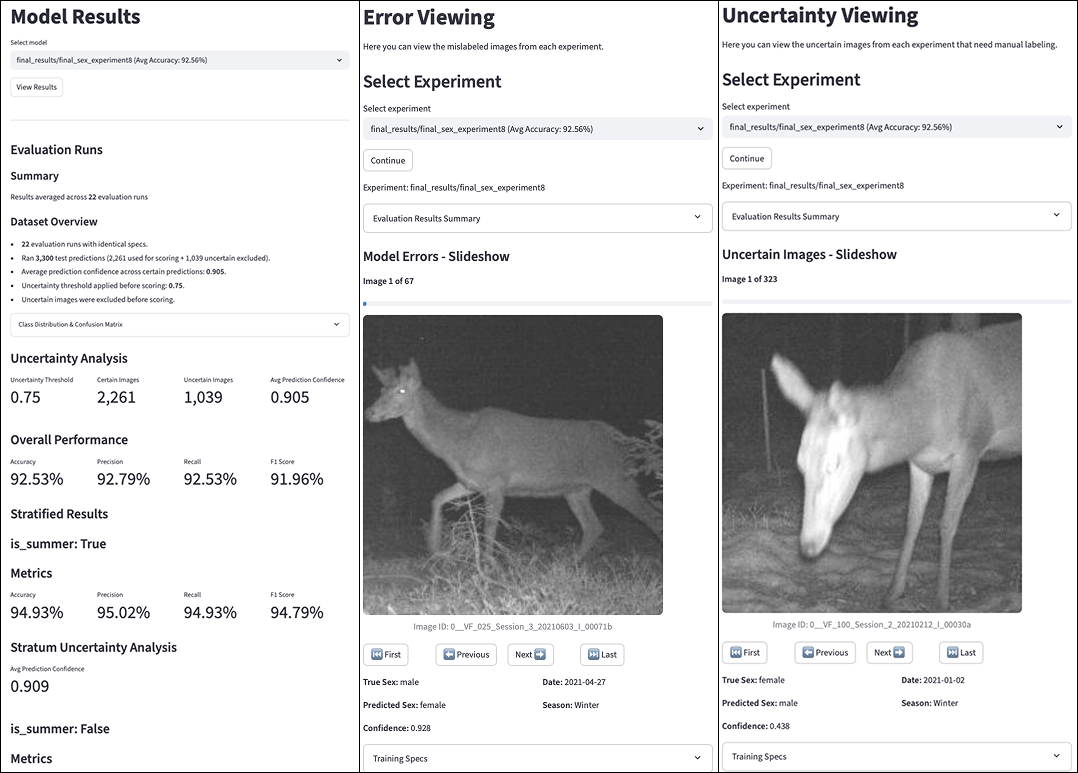}
  \caption{Model Comparison and Error/Uncertainty Viewing in Streamlit}
  \label{fig:model_comparison}
\end{figure}

The Model Results page (\autoref{fig:model_comparison}, left) summarizes the performance of each model, the number of uncertain images, and the prediction confidence, aggregated across multiple experimental runs over the same dataset to control for variation from random initializations. It also displays the class distribution in the test set and a confusion matrix of the predictions in aggregate (collapsed section in \autoref{fig:model_comparison}) and for each individual experimental run of a model (not shown here).

In the Error and Uncertainty Viewing pages (\autoref{fig:model_comparison}, middle and right, respectively), users can navigate through the images that were erroneously predicted or predicted with low confidence. This additional transparency during model selection and parameter tuning allows users to dynamically adjust their experiments based on immediate feedback on model predictions.

\section{Results}
\subsection{Data}

For the proof of concept, our dataset comprises 3,392 camera trap images of red deer (\textit{Cervus elaphus}) from the Veldenstein Forest in Northern Bavaria, Germany, collected between July 2020 and May 2022. The dataset includes 4,352 animal bounding boxes, which were manually labeled for age (adult, juvenile, yearling, unknown) and sex (male, female, unknown) by wildlife biologists at the Bavarian State Institute of Forestry (LWF).

\begin{table}[h!]
    \centering
    \caption{Distribution of true age and sex classes among high-confidence ($\geq$ 0.96, excluding bad bounding boxes) and all detections.}
    \label{tab:class-dist}
    \begin{tabular}{l l r r}
        \toprule
        Demographic & Class & CT $\geq$ 0.96 & Overall \\
        \midrule
        \multirow{3}{*}{Sex} 
            & female   & 417 (59.3\%) & 1928 (56.3\%) \\
            & male     & 262 (37.3\%) & 1351 (39.4\%) \\
            & unknown  & 24 (3.4\%) & 148 (4.3\%) \\
        \midrule
        \multirow{4}{*}{Age} 
            & adult    & 573 (77.9\%) & 2946 (74.3\%) \\
            & juvenile & 59 (8.0\%) & 517 (13.0\%) \\
            & yearling & 92 (12.5\%) & 420 (10.6\%) \\
            & unknown  & 12 (1.6\%) & 84 (2.1\%) \\ 
        \bottomrule
    \end{tabular}
\end{table}

For age and sex classification, we used a default CT of 0.96, selecting approximately 17\% of the bounding boxes. In our task, filtering out bounding boxes below this threshold changed the distribution of the class labels, especially for sex. However, we hypothesized that this change was due to filtering out poor-quality bounding boxes labeled as "unknown" rather than due to a true difference in the class label distributions. To evaluate this hypothesis, we manually coded 250 bounding boxes that had an unknown label for either age or sex. For each bounding box, we identified whether the label was due to poor image quality or missing physical characteristics.

Based on our manual coding, overall, most unknown labels were linked to poor image quality -- 82.0\% for age and 86.2\% for sex -- while the remaining 18.0\% and 13.8\%, respectively, were due to missing physical characteristics. For high-confidence detections (CT $\geq$ 0.96), the share of poor-quality images among observations labeled "unknown" decreased to 53.8\% for age and 66.7\% for sex.

This pattern indicates that high-confidence detections are less affected by poor image quality and more often reflect individuals lacking visible sex or age markers. We used the empirical distribution above to filter out the portion of bounding boxes in each unknown segment that were unknown due to poor image quality. After this filtering, the distributions of labels for age and sex between the high confidence bounding boxes and the full sample are approximately equivalent as shown in \autoref{tab:class-dist}.

\subsection{Experimental Results}
For each classification task, we conducted multiple training runs to identify high-performing parameter configurations. Initial small-scale experiments helped us determine a promising set of training parameters, which we then paired with various pretrained backbone models for both age and sex classification. After selecting the best-performing backbone for each task, we expanded the search to a broader parameter space. Across experiments, medium and strong data augmentation consistently improved performance without requiring additional training data. Although the dataset exhibits substantial class imbalance -- particularly an over-representation of adult and female deer -- data augmentation helped reduce variability between runs.

Performance evaluation used accuracy and the standard classification metrics of precision, recall, and F1-score. Each metric was computed independently for each class and then aggregated using class-frequency weighting so that classes with more samples contributed proportionally to the overall score. Due to this weighting, the accuracy and recall metrics are equivalent, but both are included for completeness. Classes with no true or predicted instances were assigned a score of zero for that metric to avoid undefined values. All experiments were repeated multiple times with the same parameter configuration to better estimate performance across random initializations and train-test data splits. We report performance metrics over all test sets, as well as the number of iterations run for each experiment.

\begin{table}[h!]
    \centering
    \caption{Performance metrics for the best-performing age and sex classification experiments.}
    \label{tab:best_age_sex_experiments}
    \begin{tabular}{l l c c}
        \toprule
        \textbf{Model} & \textbf{Metric} & \textbf{Value} & \textbf{Iterations} \\
        \midrule

        \multirow{4}{*}{\textbf{Age Classifier}}
            & Accuracy & 90.68\% & 24 \\
            & Precision & 82.22\% & 24 \\
            & Recall & 90.68\% & 24 \\
            & F1-score & 86.24\% & 24 \\

        \midrule

        \multirow{4}{*}{\textbf{Sex Classifier}}
            & Accuracy & 92.53\% & 22 \\
            & Precision & 92.79\% & 22 \\
            & Recall & 92.53\% & 22 \\
            & F1-score & 91.96\% & 22 \\
        \bottomrule
    \end{tabular}
\end{table}

For age classification, the final set of 24 evaluation runs produced an average accuracy of 90.68\%, precision of 82.22\%, recall of 90.68\%, and an F1-score of 86.24\%. Each evaluation run used an uncertainty threshold of 0.75, resulting in an average of 2,070 certain predictions and 1,530 excluded predictions across all runs. The average prediction confidence for certain predictions was 0.845.

For sex classification, 22 evaluation runs resulted in higher overall performance: 92.53\% accuracy, 92.79\% precision, 92.53\% recall, and an F1-score of 91.96\%. With the same 0.75 uncertainty threshold, the model retained an average of 2,261 certain predictions and excluded 1,039 predictions across all runs. The average prediction confidence for certain predictions was 0.905. The majority of misclassifications were cases predicted as female, which aligns with the imbalance towards female deer in the dataset.

Our results achieve a level of accuracy that is sufficient for the ecology tasks we defined given the low amount of data. In a production application, the experimental results described above would be used to select a best performing configuration of model parameters that would then be used to train a model with our tool once more data are collected. That model can then be transferred to a production server and used for inference in real time.

\section{Discussion and Conclusion}

We propose a comprehensive tool that allows domain experts to quickly train and iterate on task-specific image classification models. As a proof of concept, we applied this workflow to sex and age classification among red deer from camera trap images, but it is generally applicable to a wide range of classification tasks and image sources. To ensure usability and robustness, our experimental setup implements a constrained range of ML architectures and methods. Within these constraints, we achieve high performance in both classification tasks. The demographic estimates found with the final models can be used for practical applications such as monitoring population growth and setting hunting quotas. We believe that domain experts in a number of fields can benefit from the freedom to train and compare small task-specific ML models using our end-to-end implementation. As experts gather new data and observe the performance of their models in live scenarios, they can easily navigate through past experiments and iterate and optimize for their use case.

As with other ML pipelines, the final performance is limited by the data available and the challenge of creating a significant amount of high-quality, accurate training data for any given task. In our proof of concept, the dataset is unbalanced, with far more adult and female deer detected than any other class, but even with this disadvantage and a relatively small dataset, we were able to achieve a usable model.

As was found in previous work, camera trap images are sometimes of poor quality (extreme close-ups, only capture part of an animal, low light, etc.) such that even experts misclassify individuals \citep{bothmann_automated_2023}. Mistakes in training data lead to mistakes in a model’s self-evaluation and/or predictions. More specifically to our task, expert annotators in ecology often use context clues such as the presence of other deer in the full image to determine if an individual is a yearling or juvenile. This information is not available to the model during training as we are performing cropping based on the MD bounding boxes in the preprocessing step, so the model cannot benefit from context clues which might help a human annotator. 

We aim to further productize this framework so that it can be leveraged at scale by  researchers for new classification tasks. Additional features could be added to support more pretrained backbone models, to incorporate the full image context as a side input to the model, or produce more interpretable final models. We would also like to extend our infrastructure to accommodate more complex, computationally intensive tasks such as species classification across multiple environments.

\bibliography{bibliography}

\end{document}